\documentclass[11pt,a4paper]{article}
\usepackage[hyperref]{acl2019}
\usepackage{times}
\usepackage{latexsym}
\usepackage{graphicx}
\usepackage{color}
\usepackage{listing}
\usepackage{times}  
\usepackage{helvet}  
\usepackage{courier}  
\usepackage{url}  
\usepackage{graphicx}  
\usepackage{multirow}
\usepackage{amsthm}
\usepackage{amssymb}
\usepackage{amsmath}
\usepackage{color}
\usepackage{hyperref}
\usepackage{comment}
\usepackage{MnSymbol}
\usepackage{makecell}
\usepackage{arydshln}
\usepackage[shortlabels]{enumitem}
\usepackage{adjustbox}
\graphicspath{ {./figures/}}

\usepackage{url}

\aclfinalcopy 

\title{UR-FUNNY: A Multimodal Language Dataset for Understanding Humor}

\author{
Md Kamrul Hasan\textsuperscript{1*}, Wasifur Rahman\textsuperscript{1*}, Amir Zadeh\textsuperscript{2},  Jianyuan Zhong\textsuperscript{1},\\ \textbf{ Md Iftekhar Tanveer\textsuperscript{1}, Louis-Philippe Morency\textsuperscript{2}, Mohammed (Ehsan) Hoque\textsuperscript{1}} \\
1 - Department of Computer Science, University of Rochester, USA\\
2 - Language Technologies Institute, SCS, CMU, USA\\
\tt \{mhasan8,echowdh2\}@cs.rochester.edu, abagherz@cs.cmu.edu,\\ \tt jzhong9@u.rochester.edu, itanveer@cs.rochester.edu,\\ \tt morency@cs.cmu.edu, mehoque@cs.rochester.edu  
}
\date{}

\begin{document}
\definecolor{gg}{RGB}{45,125,45}
\definecolor{rr}{RGB}{185,45,45}

\twocolumn[{
\renewcommand\twocolumn[1][]{#1}
\maketitle 
\begin{center}
    \includegraphics[width=\linewidth]{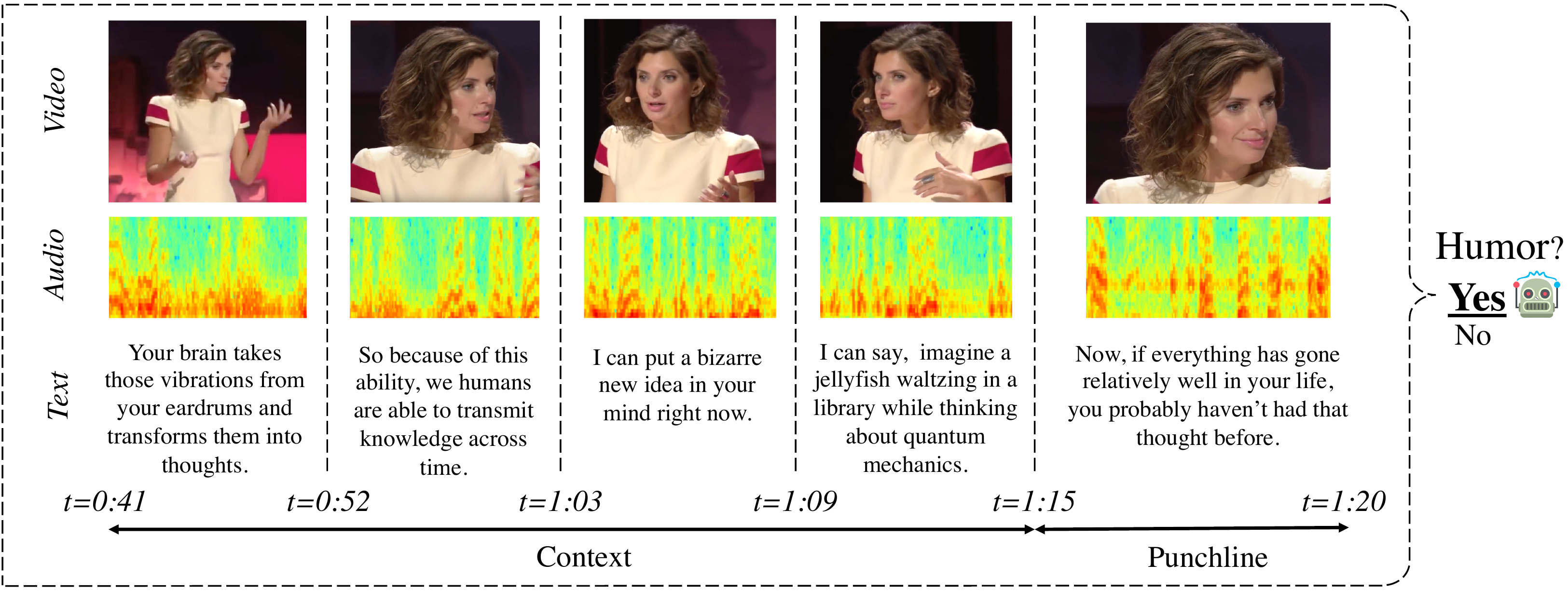}
    \captionof{figure}{\label{fig:teaser}An example of the UR-FUNNY dataset. UR-FUNNY presents a framework to study the dynamics of humor in multimodal language. Machine learning models are given a sequence of sentences with the accompanying modalities of vision and acoustic. Their goal is to detect whether or not the sequence will trigger immediate laughter by detecting whether or not the last sentence constitutes a punchline. \vspace{.7cm}}
\end{center} }]
\begin{abstract}
Humor is a unique and creative communicative behavior displayed during social interactions. It is produced in a multimodal manner, through the usage of words (text), gestures (vision) and prosodic cues (acoustic). Understanding humor from these three modalities falls within boundaries of multimodal language; a recent research trend in natural language processing that models natural language as it happens in face-to-face communication. Although humor detection is an established research area in NLP, in a multimodal context it is an understudied area. This paper presents a diverse multimodal dataset, called UR-FUNNY, to open the door to understanding multimodal language used in expressing humor. The dataset and accompanying studies, present a framework in multimodal humor detection for the natural language processing community. UR-FUNNY is publicly available for research. 
\end{abstract}

\section{Introduction}
Humor is a unique communication skill that removes barriers in conversations. Research shows that effective use of humor allows a speaker to establish rapport~\cite{stauffer1999let}, grab attention~\cite{wanzer2010explanation}, introduce a difficult concept without confusing the audience \cite{garner2005humor} and even to build trust~\cite{vartabedian1993humor}. Humor involves multimodal communicative channels including effective use of words (text), accompanying gestures (vision) and sounds (acoustic). Being able to mix and align those modalities appropriately is often unique to individuals, attributing to many different styles. Styles include gradually building up to a punchline using text, audio, video or in combination of any of them, a sudden twist to the story with an unexpected punchline~\cite{ramachandran1998neurology}, creating a discrepancy between modalities (e.g., something funny being said without any emotion, also known as dry humor), or just laughing with the speech to stimulate the audience to mirror the laughter~\cite{provine1992contagious}. 

Modeling humor using a computational framework is inherently challenging due to factors such as: 1) \textit{Idiosyncrasy}: often humorous people are also the most creative ones~\cite{hauck1972relationship}. This creativity in turn adds to the dynamic complexity of how humor is expressed in a multimodal manner. Use of words, gestures, prosodic cues and their (mis)alignments are toolkits that a creative user often experiments with. 2) \textit{Contextual Dependencies}: humor often develops through time as speakers plan for a punchline in advance. There is a gradual build up in the story with a sudden twist using a punchline \cite{ramachandran1998neurology}. Some punchlines when viewed in isolation (as illustrated in Figure \ref{fig:teaser}) may not appear funny. The humor stems from the prior build up, cross-referencing multiple sources, and its delivery.  Therefore, a full understanding of humor requires analyzing the context of the punchline. 

Understanding the unique dependencies across modalities and its impact on humor require knowledge from multimodal language; a recent research trend in the field of natural language processing \cite{zadeh2018proceedings}. Studies in this area aim to explain natural language from three modalities of text, vision and acoustic. In this paper, alongside computational descriptors for text, gestures such as smile or vocal properties such as loudness are measured and put together in a multimodal framework to define humor recognition as a multimodal task.


The main contribution of this paper to the NLP community is introducing the first multimodal language (including text, vision and acoustic modalities) dataset of humor detection named ``UR-FUNNY". This dataset opens the door to understanding and modeling humor in a multimodal framework. The studies in this paper present performance baselines for this task and demonstrate the impact of using all three modalities together for humor modeling. 

\section{Background}
The dataset and experiments in this paper are connected to the following areas:

\noindent \textbf{Humor Analysis:} Humor analysis has been among active areas of research in both natural language processing and affective computing. Notable datasets in this area include ``16000 One-Liners'' \cite{mihalcea2005making}, ``Pun of the Day'' \cite{yang2015humor}, ``PTT Jokes'' ~\cite{chen2018humor}, ``Ted Laughter''~\cite{chen2017predicting}, and ``Big Bang Theory'' ~\cite{bertero2016deep}. The above datasets have studied humor from different perspectives. For example, ``16000 One-Liner'' and ``Pun of the Day'' focus on joke detection (joke vs. not joke binary task), while ``Ted Laughter'' focuses on punchline detection (whether or not punchline triggers laughter). Similar to ``Ted Laughter'', UR-FUNNY focuses on punchline detection. Furthermore, punchline is accompanied by context sentences to properly model the build up of humor. Unlike previous datasets where negative samples were drawn from a different domain, UR-FUNNY uses a challenging negative sampling case where samples are drawn from the same videos. Furthermore, UR-FUNNY is the only humor detection dataset which incorporates all three modalities of text, vision and audio. Table \ref{table:comparison} shows a comparison between previously proposed datasets and UR-FUNNY dataset.

From modeling aspect, humor detection is done using hand-crafted and non-neural models ~\cite{yang2015humor}, neural based RNN and CNN models for detecting humor in Yelp ~\cite{de2017humor} and TED talks ~\cite{chen2017predicting}. Newer approaches have used ~\cite{chen2018humor} highway networks ``16000 One-Liner'' and ``Pun of the Day'' datasets. There have been very few attempts at using extra modalities alongside language for detecting humor, mostly limited to adding simple audio features ~\cite{rakov2013sure, bertero2016deep}. Furthermore, these attempts have been restricted to certain topics and domains (such as ``Big Bang Theory'' TV show \cite{bertero2016deep}).


\begin{table}[t!]
\begin{center}
\small
\setlength{\tabcolsep}{2.9pt}
\begin{tabular}{l:c:c:c:c:c}
\Xhline{3\arrayrulewidth}
 Dataset & \#Pos & \#Neg & Mod & type & \#spk \\\hline 
16000 One-Liners & 16000 & 16000 & \{\textit{l}\} & joke &-\\
Pun of the Day & 2423 & 2423 & \{\textit{l}\} & pun & -\\
PTT Jokes & 1425 & 2551 & \{\textit{l}\} & political & -\\
Ted Laughter & 4726 & 4726 & \{\textit{l}\} &  speech & 1192 \\
Big Bang Theory & 18691 & 24981 & \{\textit{l,a}\} & tv show & $<$50 \\
\textbf{UR-Funny} & 8257 & 8257 & \{\textit{l,a,v}\} & speech&1741\\
\Xhline{3\arrayrulewidth}
\end{tabular}
\end{center}
\caption{Comparison between UR-FUNNY and notable humor detection datasets in the NLP community. Here, `pos', 'neg' , `mod' and `spk' denote  positive, negative, modalities and speaker respectively.}
\label{table:comparison}
\end{table}


\noindent \textbf{Multimodal Language Analysis:} Studying natural language from modalities of text, vision and acoustic is a recent research trend in natural language processing \cite{zadeh2018proceedings}. Notable works in this area present novel multimodal neural architectures \cite{wang2018words,pham2018found,hazarika2018conversational,poria2017multi,zadeh2017tensor}, multimodal fusion approaches \cite{liang2018multimodal,tsai2018learning,liu2018efficient,zadeh2018memory,barezi2018modality} as well as resources \cite{poria2018meld,zadeh2018multimodal,zadeh2016mosi,park2014computational,rosas2013multimodal,wollmer2013youtube}. Multimodal language datasets mostly target multimodal sentiment analysis \cite{poria2018multimodal}, emotion recognition \cite{zadeh2018multimodal,busso2008iemocap}, and personality traits recognition \cite{park2014computational}. UR-FUNNY dataset is similar to the above datasets in diversity (speakers and topics) and size, with the main task of humor detection. Beyond the scope of multimodal language analysis, the dataset and studies in this paper have similarities to other applications in multimodal machine learning such language and vision studies, robotics, image captioning, and media description \cite{baltruvsaitis2019multimodal}.


\section{UR-FUNNY Dataset}
In this section we present the UR-FUNNY dataset. We first discuss the data acquisition process, and subsequently present statistics of the dataset as well as multimodal feature extraction and validation.

\subsection{Data Acquisition}
A suitable dataset for the task of multimodal humor detection should be diverse in a) \textit{speakers}: modeling the idiosyncratic expressions of humor may require a dataset with large number of speakers, and b) \textit{topics}: different topics exhibit different styles of humor as the context and punchline can be entirely different from one topic to another. 

TED talks~\footnote{Videos on \url{www.ted.com} are publicly available for download.} are among the most diverse idea sharing channels, in both speakers and topics. Speakers from various backgrounds, ethnic groups and cultures present their thoughts through a widely popular channel \footnote{More than 12 million subscribers on YouTube \url{https://www.youtube.com/user/TEDtalksDirector}}. The topics of these presentations are diverse; from scientific discoveries to everyday ordinary events. As a result of diversity in speakers and topics, TED talks span across a broad spectrum of humor. Therefore, this platform presents a unique resource for studying the dynamics of humor in a multimodal setup. 

TED videos include manual transcripts and audience markers. Transcriptions are highly reliable, which in turn allow for aligning the text and audio. This property makes TED talks a unique resource for newest continuous fusion trends \cite{chen2017multimodal}. Transcriptions also include reliably annotated markers for audience behavior. Specifically, the ``laughter'' marker has been used in NLP studies as an indicator of humor \cite{chen2017predicting}. Previous studies have identified the importance of both punchline and context in understanding and modeling the humor. In a humorous scenario, context is the gradual build up of a story and punchline is a sudden twist to the story which causes laughter \cite{ramachandran1998neurology}. Using the provided laughter marker, the sentence immediately before the marker is considered as the punchline and the sentences prior to punchline (but after previous laughter marker) are considered context. 

\begin{figure*}
\includegraphics[width=1\linewidth]{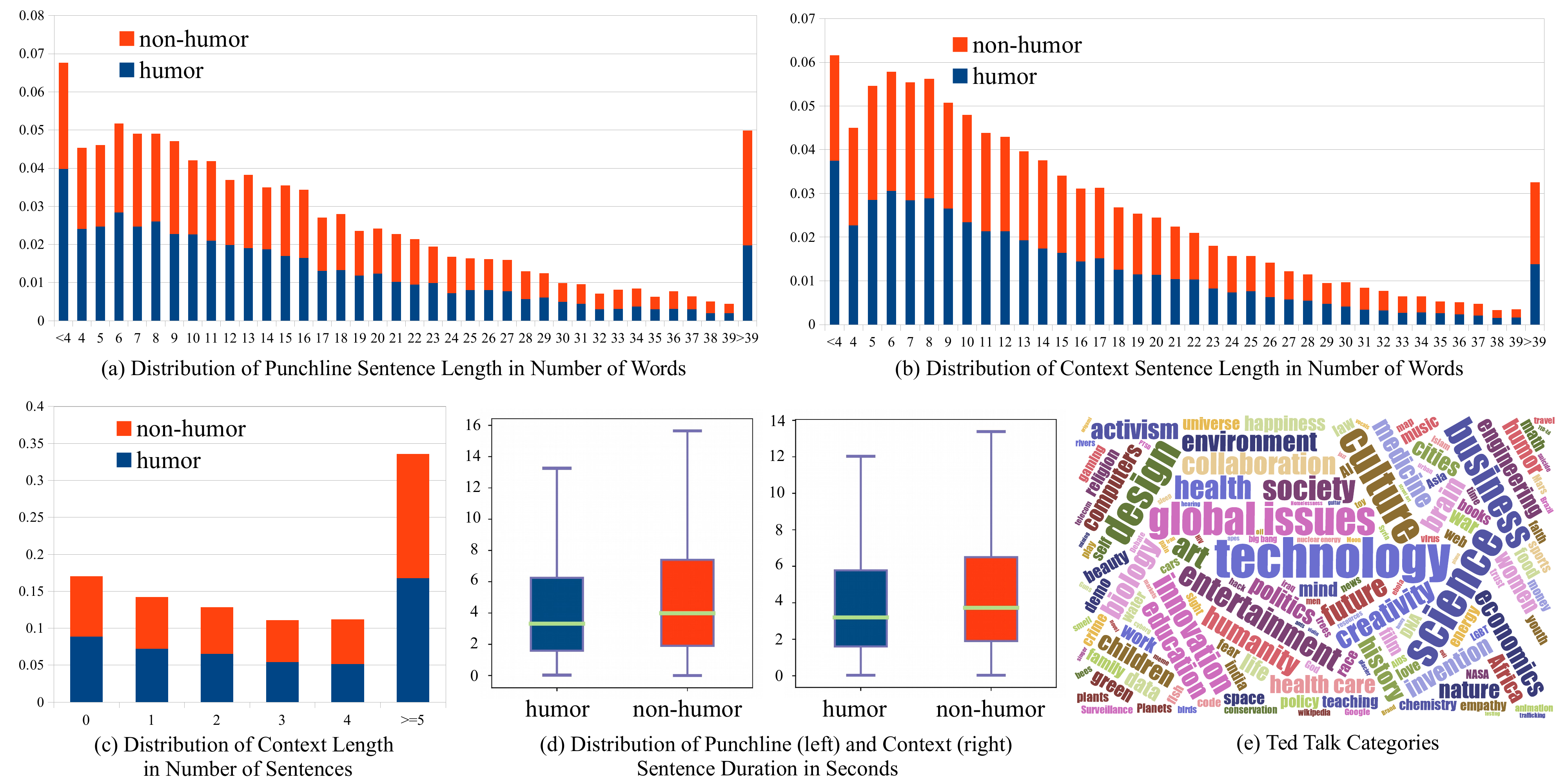}
\caption{Overview of UR-FUNNY dataset statistics. (a) the distribution of punchline sentence length for humor and non-humor cases. (b) the distribution of context sentence length for humor and non-humor cases. (c) distribution of the number of sentences in the context. (d) distribution of the duration (in seconds) of punchline and context sentences. (e) topics of the videos in UR-FUNNY dataset. Best viewed in zoomed and color.}
\label{humordata}
\end{figure*}

We collect $1866$ videos as well as their transcripts from TED portal. These $1866$ videos are chosen from $1741$ different speakers and across $417$ topics. The laughter markup is used to filter out $8257$ humorous punchlines from the transcripts~\cite{chen2017predicting}. The context is extracted from the prior sentences to the punchline (until the previous humor instances or the beginning of video is reached). Using a similar approach, $8257$ negative samples are chosen at random intervals where the last sentence is not immediately followed by a laughter marker. The last sentence is assumed a punchline and similar to the positive instances, the context is chosen. This negative sampling uses sentences from the same distribution, as opposed to datasets which use sentences from other distributions or domains as negative sample ~\cite{yang2015humor,mihalcea2005making}. After this negative sampling, there is a homogeneous $50\%$ split in the dataset between positive and negative examples. 

Using forced alignment, we mark the beginning and end of each sentence in the video as well as words and phonemes in the sentences \cite{yuan2008speaker}. Therefore, an alignment is established between text, audio and video. Utilizing this alignment, the timing of punchline as well as context is extracted for all instances in the dataset.



\begin{table}[t!]
\begin{center}
\small
\setlength{\tabcolsep}{2pt}
\begin{tabular}{l:c}
\Xhline{3\arrayrulewidth}
\textbf{General} \\
total \#videos & 1866 \\
total duration in hour & 90.23 \\
total \#distinct speakers & 1741 \\
total \#distinct topics & 417 \\
total \#humor instances & 8257 \\
total \#non-humor instances & 8257 \\
total \#words & 965573 \\ 
total \#unique words & 32995 \\ 
total \#sentences & 63727 \\
avg length of sentences in words & 15.15 \\ 
avg duration of sentences (s) & 4.64 \\ 
\hline
\textbf{Punchline} \\
\#sentences in punchline & 1 \\
avg \#words in punchline & 16.14\\
avg \#words in humorous punchline & 15.17\\
avg \#words in non-humorous punchline &17.10\\
avg duration of punchline (s) & 4.97 \\
avg duration of humorous punchline (s) & 4.58  \\
avg duration of non-humorous punchline (s) &5.36 \\
\hline
\textbf{Context} \\
avg total \#words in context &42.33\\
avg \#words in context sentences &14.80\\
avg \#sentences in context & 2.86\\
avg \#sentences in humorous context & 2.82\\
avg \#sentences in non-humorous context & 2.90\\
avg duration of context (s) & 14.7 \\
avg duration of humorous context (s) & 13.79 \\
avg duration of non-humorous context (s) & 15.62 \\
avg duration of context sentences (s) & 4.25 \\
avg duration of humorous context sentences (s) & 4.79 \\
avg duration of non-humorous context sentences (s) & 4.52 \\
\Xhline{3\arrayrulewidth}
\end{tabular}
\end{center}
\caption{\label{summary_table}Summary of the UR-FUNNY dataset statistics. Here, `\#' denotes number, `avg' denotes average and `s' denotes seconds}
\end{table}

\subsection{Dataset Statistics}\label{sec:dataset} 
The high level statistics of UR-FUNNY dataset are presented in Table \ref{summary_table}. Total duration of the entire dataset is $90.23$ hours. There are a total of $1741$ distinct speakers and a total of $417$ distinct topics in the UR-FUNNY dataset. Figure \ref{humordata}.e shows the word cloud of the topics based on log-frequency of the topic. The top most five frequent topics are technology, science, culture, global issues and design \footnote{Metadata collected from \url{www.ted.com}}. There are in total $16514$ video segments of humor and not humor instances (equal splits of $8257$). The average duration of each data instance is $19.67$ seconds, with context an average of $14.7$ and punchline with an average of $4.97$ seconds. The average number of words in punchline is $16.14$ and the average number of words in context sentences is $14.80$. 

Figure \ref{humordata} shows an overview for some of the important statistics of UR-FUNNY dataset. Figure \ref{humordata}.a demonstrates the distribution of punchline for humor and non-humor cases based on number of words. There is no clear distinction between humor and non-humor punchlines as both follow similar distribution. Similarly, Figure \ref{humordata}.b shows the distribution of number of words per context sentence. Both humor and non-humor context sentences follow the same distribution. Majority $(\geq 90\%)$ of punchlines have length less than $32$. In terms of number of seconds, Figure \ref{humordata}.d shows the distribution of punchline and context sentence length in terms of seconds. Figure \ref{humordata}.c demonstrates the distribution of number of context sentences per humor and non-humor data instances. Number of context sentences per humor and non-humor case is also roughly the same. The statistics in Figure \ref{humordata} show that there is no trivial or degenerate distinctions between humor and non-humor cases. Therefore, classification of humor versus non-humor cases cannot be done based on simple measures (such as number of words); it requires understanding the content of sentences.


\begin{table}[t!]
\begin{center}
\small
\setlength{\tabcolsep}{5pt}
\begin{tabular}{l: c : c : c}
\Xhline{3\arrayrulewidth}
 & Train & Val & Test \\ \hline
\#humor instances & 5306 & 1313 & 1638 \\
\#not humor instances & 5292 & 1313 & 1652 \\
\#videos used & 1166 & 300 & 400 \\
\#speakers & 1059 & 294 & 388\\
avg \#words in punchline & 15.81 & 16.94 & 16.55 \\
avg \#words in context & 41.69 & 42.86 & 43.94 \\
avg \#sentences in context & 2.84 & 2.81 & 2.95 \\
punchline avg duration(second) & 4.85 & 5.25 & 5.15 \\
context avg duration(second)  &  14.39 & 14.91 & 15.54 \\
\Xhline{3\arrayrulewidth}
\end{tabular}
\end{center}
\caption{Statistics of train, validation \& test folds of UR-FUNNY dataset. Here, `avg' denotes average and `\#' denotes number. }
\label{summary2}
\end{table}

Table \ref{summary2} shows the standard train, validation and test folds of the UR-FUNNY dataset. These folds share no speaker with each other - hence standard folds are speaker independent \cite{zadeh2016mosi}. This minimizes the chance of overfitting to identity of the speakers or their communication patterns. 

\subsection{Extracted Features}
For each modality, the extracted features are as follows: \newline

\noindent \textbf{Language:} Glove word embeddings \cite{pennington2014glove} are used as pre-trained word vectors for the text features. P2FA forced alignment model \cite{yuan2008speaker} is used to align the text and audio on phoneme level. From the force alignment, we extract the timing annotations of context and punchline on word level. Then, the acoustic and visual cues are aligned on word level by interpolation \cite{chen2017multimodal}. \newline
  
\noindent \textbf{Acoustic:} COVAREP software \cite{degottex2014covarep} is used to extraction acoustic features at the rate of 30 frame/sec. We extract following 81 features: fundamental frequency (F0), Voiced/Unvoiced segmenting features (VUV) \cite{drugman2011joint}, normalized amplitude quotient (NAQ), quasi open quotient (QOQ) \cite{kane2013wavelet}, glottal source parameters (H1H2, Rd,Rd conf) \cite{drugman2012detection,alku2002normalized,alku1997parabolic}, parabolic spectral parameter (PSP), maxima dispersion quotient (MDQ), spectral tilt/slope of wavelet responses (peak/slope), Mel cepstral coefficient (MCEP 0-24), harmonic model and phase distortion mean (HMPDM 0-24) and deviations (HMPDD 0-12), and the first 3 formants. These acoustic features are related to emotions and tone of speech. \newline

\noindent \textbf{Visual:} OpenFace facial behavioral analysis tool \cite{baltruvsaitis2016openface} is used to extract the facial expression features at the rate of 30 frame/sec. We extract all facial Action Units (AU) features based on the Facial Action Coding System (FACS) \cite{ekman1997face}. Rigid and non-rigid facial shape parameters are also extracted \cite{baltruvsaitis2016openface}. We observed that the camera angle and position changes frequently during TED presentations. However, for the majority of time, the camera stays focused on the presenter. Due to the volatile camera work, the only consistently available source of visual information was the speaker's face. \newline

UR-FUNNY dataset is publicly available for download alongside all the extracted features. 
\section{Multimodal Humor Detection}

In this section, we first outline the problem formulation for performing binary multimodal humor detection on UR-FUNNY dataset. We then proceed to study the UR-FUNNY dataset through the lens of a contextualized extension of Memory Fusion Network (MFN) \cite{zadeh2018memory} - a state-of-the-art model in multimodal language. 

\begin{figure}
\begin{center}
\includegraphics[width=\linewidth]{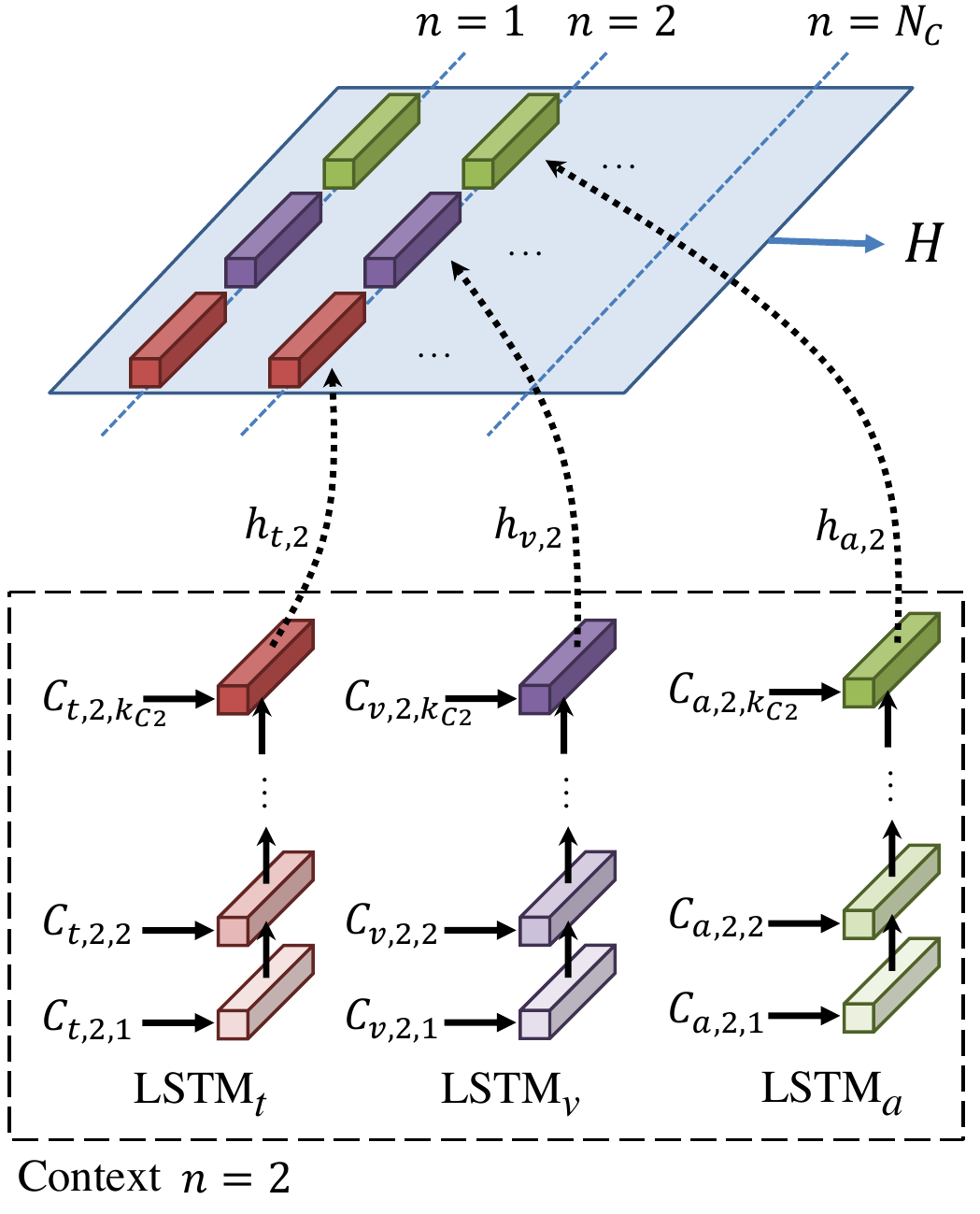}
\caption{\label{fig:unimodal}The structure of Unimodal Context Network as outlined in Section \ref{subsec:unimodal_net}. For demonstration purpose, we show the case for $n=2$ (second context sentence). After $n=N_C$, the output $H$ (outlined by blue) is complete. Best viewed in color.}
\end{center}
\end{figure}

\subsection{Problem Formulation}
UR-FUNNY dataset is a multimodal dataset with three modalities of text, vision and acoustic. We denote the set of these modalities as $M=\{t,v,a\}$. Each of the modalities come in a sequential form. We assume word-level alignment between modalities \cite{yuan2008speaker}. Since frequency of the text modality is less than vision and acoustic (i.e. vision and acoustic have higher sampling rate), we use expected visual and acoustic descriptors for each word~\cite{chen2017multimodal}. After this process, each modality has the same sequence length (each word has a single vision and acoustic vector accompanied with it). 

Each data sample in the UR-FUNNY can be described as a triplet  $(l,P,C)$ with $l$ being a binary label for humor or non-humor. $P$ is the punchline and $C$ is the context. Both punchline and context have multiple modalities $P=\{P_m; m \in M\}$, $C=\{C_m; m \in M\}$. If there are $N_C$ context sentences accompanying the punchline, then $C_m = [C_{m,1},C_{m,2},\dots, C_{m,N_C}]$ - simply context sentences start from first sentence to the last ($N_C$) sentence. $K_P$ is the number of words in the punchline and $K_{Cn}|_{n=1}^{N_C}$ is the number of words in each of the context sentences respectively. As examples of this notation, $P_{m,k}$ refers to the $k$th entry in the modality $m$ of the punchline. $C_{m,n,k}$ refers to the $k$th entry in the modality $m$ of the $n$th context. 

Models developed on UR-FUNNY dataset are trained on triplets of $(l,P,C)$. During testing only a tuple $(P,C)$ is given to predict the $l$. $l$ is the label for laughter, specifically whether or not the inputs $P,C$ are likely to trigger a laughter.

\subsection{Contextual Memory Fusion Baseline} \label{subsec:cmfn}

Memory Fusion Network (MFN) is among the state-of-the-art models for several multimodal datasets \cite{zadeh2018memory}. We devise an extension of the MFN model, named Contextual Memory Fusion Network~\footnote{Code available through hidden-for-blind-review.}(C-MFN), as a baseline for humor detection on UR-FUNNY dataset. This is done by introducing two components to allow the involvement of context in the MFN model: 1) \textit{Unimodal Context Network}, where information from each modality is encoded using $M$ Long-short Term Memories (LSTM), 2) \textit{Multimodal Context Network}, where unimodal context information are fused (using self-attention) to extract the multimodal context information. We discuss the components of the C-MFN model in the continuation of this section. 

\subsubsection{Unimodal Context Network}\label{subsec:unimodal_net}

To model the context, we first model each modality within the context. Unimodal Context Network (Figure \ref{fig:unimodal}) consists of $M$ LSTMs, one for each modality $m \in M$ denoted as LSTM$_m$. For each context sentence $n$ of each modality $m \in M$, LSTM$_m$ is used to encode the information into a single vector $h_{m,n}$. This single vector is the last output of the LSTM$_m$ over $C_{m,n}$ as input. The recurrence step for each LSTM is the utterance of each word (due to word-level alignment vision and acoustic modalities also follow this time-step). The output of the Unimodal Context Network is the set $H=\{h_{m,n};m \in M, 1 \leq n < N_C\}$. 

\subsubsection{Multimodal Context Network}\label{subsec:multimodal_net}

\begin{figure}
\begin{center}
\includegraphics[width=\linewidth]{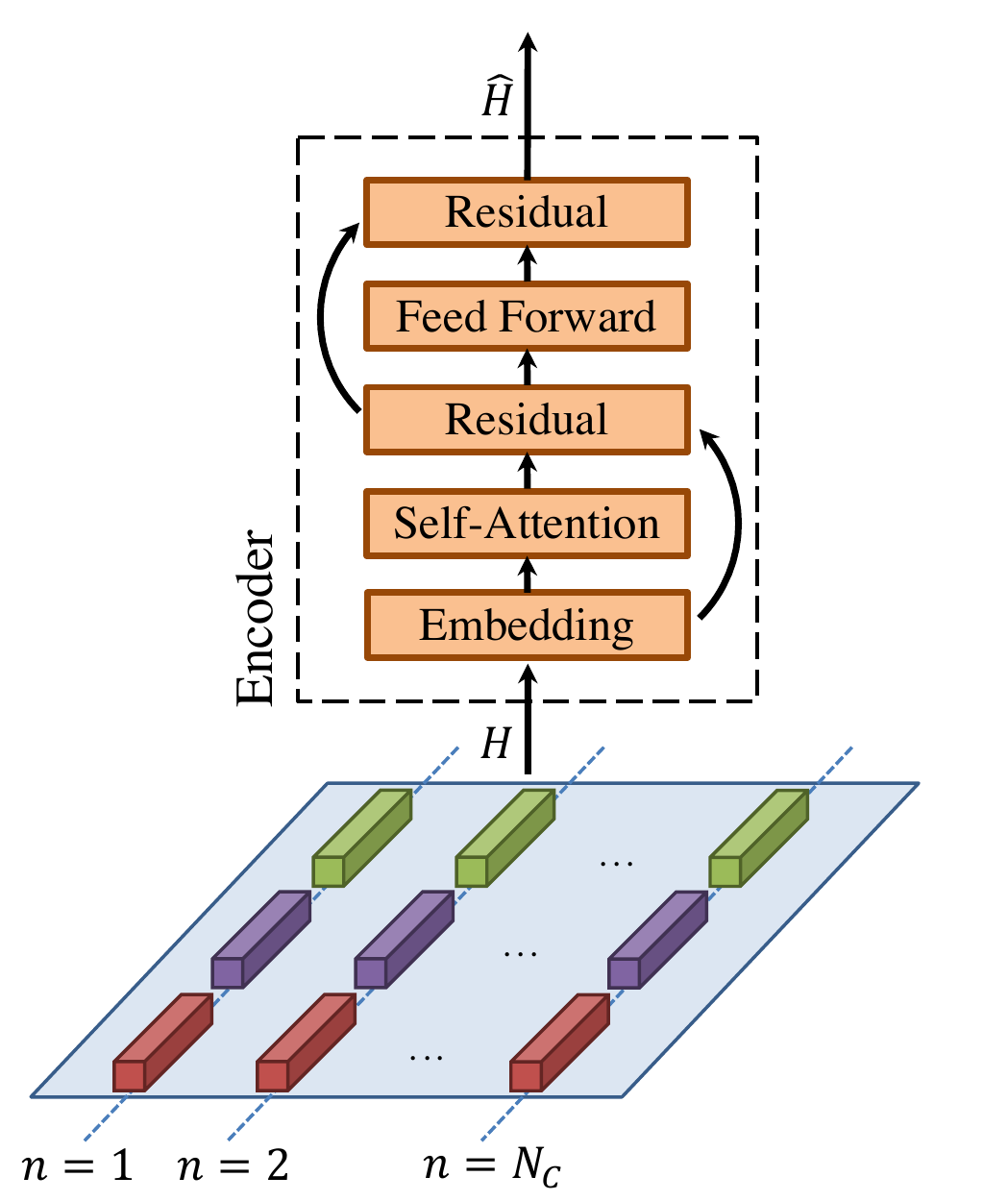}
\caption{\label{fig:multimodal}The structure of Multimodal Context Network as outlined in Section \ref{subsec:multimodal_net}. The output $H$ of the Unimodal Context Network is connected to an encoder module to get the multimodal output $\hat{H}$. For the details of components outlined in orange please refer to the authors' original paper. \cite{vaswani2017attention}. Best viewed in color. } 
\end{center}
\end{figure}

Multimodal Context Network (Figure \ref{fig:multimodal}) learns a multimodal representation of the context based on the output $H$ of the Unimodal Context Network. Sentences and modalities in the context can form complex asynchronous spatio-temporal relations. For example, during the gradual buildup of the context, the speaker's facial expression may be impacted due to an arbitrary previously uttered sentence. Transformers \cite{vaswani2017attention} are a family of neural models that specialize in finding various temporal relations between their inputs through self-attention. By concatenating representations $h_{m \in M, n}$ (i.e. for all $M$ modalities of the $n$th context), self-attention model can be applied to find asynchronous spatio-temporal relations in the context. We use an encoder with $6$ intermediate layers to derive a multimodal representation $\hat{H}$ conditioned on $H$. $\hat{H}$ is also spatio-temporal (as produced output of encoders in a transformer are). The output of Multimodal Context Network is the output $\hat{H}$ of the encoder. 

\subsubsection{Memory Fusion Network (MFN)}\label{subsec:mfn}

\begin{figure}
\begin{center}
\includegraphics[width=\linewidth]{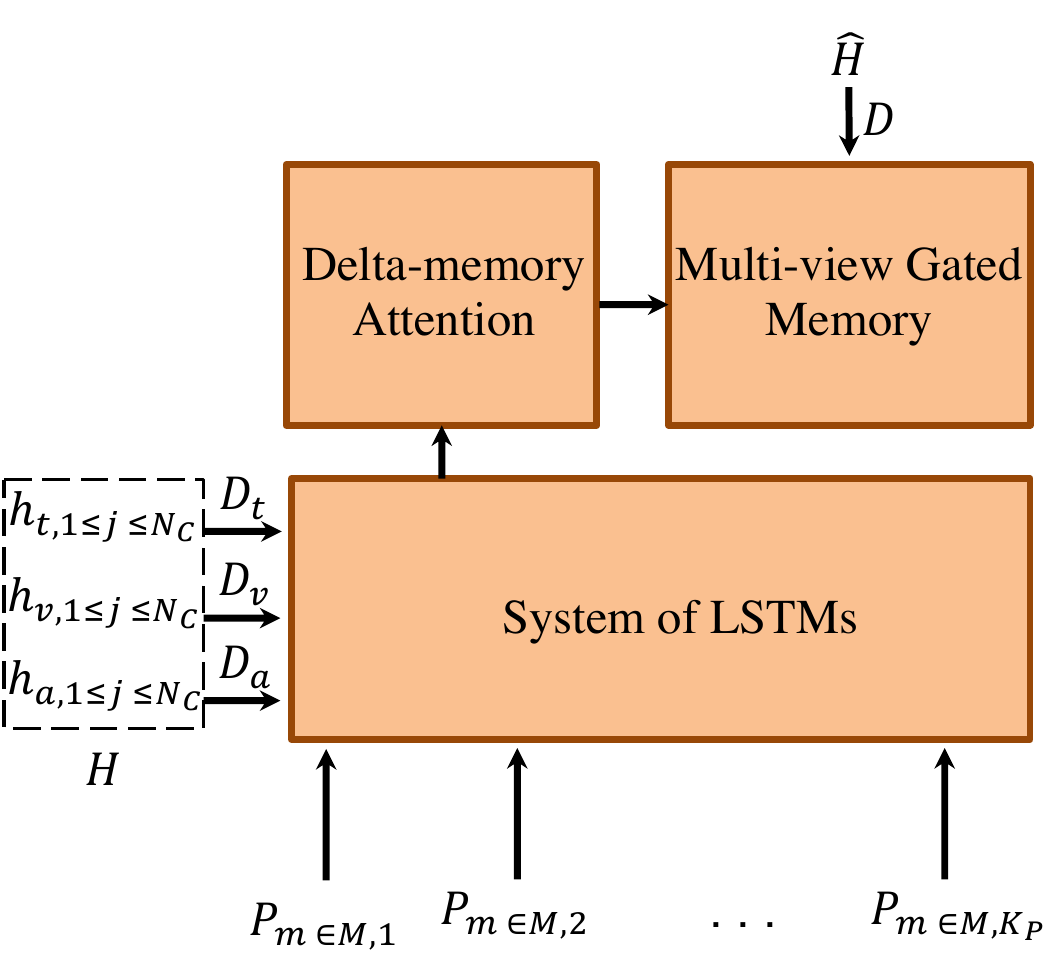}
\caption{\label{fig:mfn}The initialization and recurrence process of Memory Fusion Network (MFN). The outputs of Unimodal and Multimodal Context Networks ($H$ and $\hat{H}$) are used initializing the MFN neural components. For the details of components outlined in orange please refer to the authors' original paper \cite{zadeh2018memory}. Best viewed in color.} 
\end{center}
\end{figure}

After learning unimodal ($H$) and multimodal ($\hat{H}$) representations of context, we use a Memory Fusion Network (MFN) to model the punchline (Figure \ref{fig:mfn}). MFN contains 2 types of memories: a System of LSTMs with $M$ unimodal memories to model each modality in punchline, and a Multi-view Gated Memory which stores multimodal information. We use a simple trick to combine the Context Networks (Unimodal and Multimodal) with the MFN: we initialize the memories in the MFN using the outputs $H$ (unimodal representation) and $\hat{H}$ (multimodal representation). For System of LSTMs, this is done by initializing the LSTM cell state of modality $m$ with $\mathcal{D}_m(h_{m,1 \leq n< N_C})$. $\mathcal{D}_m$ is a fully connected neural network that maps the information from $h_{m,1 \geq j \geq N_C}$ ($m$th modality in context) to the cell state of the $m$th LSTM in the System of LSTMs. The Multi-view Gated Memory is initialized based on a non-linear projection $\mathcal{D}(\hat{H})$ where $\mathcal{D}$ is a fully connected neural network. Similar to context where modalities are aligned at word level, punchline is also aligned the same way. Therefore a word-level implementation of the MFN is used, where a word and accompanying vision and acoustic descriptors are used as input to the System of LSTMs  at each time-step. The Multi-view Gated Memory is updated iteratively at every recurrence of the System of LSTMs using a Delta-memory Attention Network. 

The final prediction of humor is conditioned on the last state of the System of LSTMs and Multi-view Gated Memory using an affine mapping with Sigmoid activation. 



\section{Experiments}\label{sec:experiments}

In the experiments of this paper, our goal is to establish a performance baseline for the UR-FUNNY dataset. Furthermore, we aim to understand the role of context and punchline, as well as role of individual modalities in the task of humor detection. For all the experiments, we use the proposed contextual extension of Memory Fusion Network (MFN), called C-MFN (Section \ref{subsec:cmfn}). Aside the proposed \textbf{C-MFN} model, the following variants are also studied: \newline

\noindent \textbf{C-MFN (P):} This variant of the C-MFN uses only punchline with no contextual information. Essentially, this is equivalent to a MFN model since initialization trick is not used. \newline

\noindent \textbf{C-MFN (C):} This variant of the C-MFN uses only contextual information without punchline. Essentially, this is equivalent to removing the MFN and directly conditioning the humor prediction on the Unimodal and Multimodal Context Network outputs (Sigmoid activated neuron after applying $D_M; m \in M$ on $H$ and $D$ on $\hat{H}$). \newline

The above variants of the C-MFN allow for studying the importance of punchline and context in modeling humor. Furthermore, we compare the performance of the C-MFN variants in the following scenarios: \textbf{(T)} a only text modality is used without vision and acoustic, \textbf{(T+V)} text and vision modalities are used without acoustic, \textbf{(T+A)} text and acoustic modalities are used without vision, \textbf{(A+V)} only vision and acoustic modalities are used, \textbf{(T+A+V)} all modalities are used together. 

We compare the performance of C-MFN variants across the above scenarios. This allows for understanding the role of context and punchline in humor detection, as well as the importance of different modalities. All the models for our experiments are trained using categorical cross-entropy. This measure is calculated between the output of the model and ground-truth labels.

\section{ Results and Discussion} 

\begin{table}[t!]
\begin{center}
\small
\setlength{\tabcolsep}{5.5pt}
\begin{tabular}{l:c:c:c:c:c}
\Xhline{3\arrayrulewidth}
Modality & T & A+V  & T+A & T+V & T+A+V\\
\hline
C-MFN (P) & 62.85 & 53.3  & 63.28 & 63.22 & 64.47 \\ 
C-MFN (C) & 57.96 & 50.23  & 57.78 & 57.99 & 58.45 \\
C-MFN & 64.44 & 57.99  & 64.47 & 64.22 & 65.23\\
\Xhline{3\arrayrulewidth}
\end{tabular}
\end{center}
\caption{Binary accuracy for different variants of C-MFN and training scenarios outlined in Section \ref{sec:experiments}. The best performance is achieved using all three modalities of text (T), vision (V) and acoustic (A).}
\label{table:humor_score}
\end{table}

The results of our experiments are presented in Table \ref{table:humor_score}. Results demonstrate that both context and punchline information are important as C-MFN outperforms C-MFN (P) and C-MFN (C) models. Punchline is the most important component for detecting humor as the performance of C-MFN (P) is significantly higher than C-MFN (C). 

Models that use all modalities (T+A+V) outperform models that use only one or two modalities (T, T+A, T+V, A+V). Between text (T) and nonverbal behaviors (A+V), text shows to be the most important modality. Most of the cases, both modalities of vision and acoustic improve the performance of text alone (T+V, T+A). 

Based on the above observations, each neural component of the C-MFN model is useful in improving the prediction of humor. The results also  indicate that modeling humor from a multimodal perspective yields successful results. 

The human performance~\footnote{This is calculated by averaging the performance of two annotators over a shuffled set of $100$ humor and $100$ non-humor cases. The annotators are given the same input as the machine learning models (similar context and punchline). The annotators agree $84\%$ of times. } on the UR-FUNNY dataset is $82.5\%$. 

The results from Table \ref{table:humor_score} demonstrate that while a state-of-the-art model can achieve a reasonable level of success in modeling humor, there is still a large gap between human-level performance with state of the art. Therefore, UR-FUNNY dataset presents new challenges to the field of NLP, specifically research areas of humor detection and multimodal language analysis.

\section{Conclusion}

In this paper, we presented a new multimodal dataset for humor detection called UR-FUNNY. This dataset is the first of its kind in the NLP community. Humor detection is done from the perspective of predicting laughter - similar to ~\cite{chen2017predicting}. UR-FUNNY is diverse in both speakers and topics. It contains three modalities of text, vision and acoustic. We study this dataset through the lens of a Contextualized Memory Fusion Network (C-MFN). Results of our experiments indicate that humor can be better modeled if all three modalities are used together. Furthermore, both context and punchline are important in understanding humor. The dataset and the accompanying experiments will be made publicly available.

\bibliography{acl2019}
\bibliographystyle{acl_natbib}

\appendix

\section{Hyperparameter Space Search}
In this appendix we present the hyperparameter space explored for C-MFN model. 
\begin{itemize}

    \item \textbf{Uni-modal Context Network:}
    \begin{enumerate}
        \item This module has three LSTMs. Hidden size for them was chosen randomly from:
        \begin{itemize}
            \item For LSTM$_l$:$[32,64,88,128,156,256]$
            \item For LSMT$_a$:$[8,16,32,48,64,80]$
            \item For LSTM$_v$:$[8,16,32,48,64,80]$
        \end{itemize}

    \end{enumerate}
    \item \textbf{Multimodal Context Network}:
    \begin{enumerate}
        \item We use the optimal configurations as described in \cite{vaswani2017attention} and implemented in \cite{jadore801120}. Some of the main configurations are:
        \begin{itemize}
            \item d\_model(output dimension of Encoder):512,
            \item d\_k(dimension of key):64,
            \item d\_v(dimension of value):64,
            \item n\_head(number of heads used in multi-headed attention):8,
            \item n\_layers(number of layers used in Encoder):6,
            \item n\_warmup\_steps:4000,
            \item dropout:0.1
        \end{itemize}
         \item To regularize the output of $D(\Tilde{H})$, we randomly choose a dropout rate from $[0.0,0.2,0.5,0.1]$
        \item To regularize the output $D_m(H)$, we use a dropout probability randomly from: 
        \begin{itemize}
            \item For m=l:$[0.0,0.1, 0.2,0.5]$
            \item For m=a:$[0.0,0.2,0.5,0.1]$
            \item For m=v:$[0.0,0.2,0.5,0.1]$
        \end{itemize}
    \end{enumerate}
    
    \item \textbf{Memory Fusion Network(MFN)}:
    \begin{enumerate}
       \item \textbf{System of LSTMs:} Hidden size of LSTM$_m$, $m \in [l,a,v]$ was randomly chosen from:
       \begin{itemize}
           \item For LSTM$_l ,[32,64,88,128,156,256]$
           \item For LSTM$_a ,[8,16,32,48,64,80]$
           \item For LSTM$_v ,[8,16,32,48,64,80]$
       \end{itemize}
       
       \item \textbf{Delta Memory Attention:}This section has two affine transformation, we call them NN1 and NN2.
       \begin{itemize}
           \item  The projection shape of NN1 is chosen randomly from $[32,64,128,256]$ and that output goes through a dropout layer whose dropout rate is chosen randomly from $[0.0,0.2,0.5,0.7]$
           \item Similarly, the projection shape of NN2 is chosen randomly from $[32,64,128,256]$ followed by a dropout layer whose dropout rate is chosen randomly from $[0.0,0.2,0.5,0.7]$.
       \end{itemize}
      
      \item \textbf{Multi-view gated memory} also has two affine transformation denoted here as Gamma1 and Gamma2.
      \begin{itemize}
          \item Gamma1 first does a projection of shape chosen randomly from
          $[32,64,128,256]$ followed by a dropout whose rate is randomly chosen from $[0.0,0.2,0.5,0.7]$.
          \item Gamma2 first does a projection and then a dropout. The projection shape is chosen randomly from
          $[32,64,128,256]$ and dropout rate is chosen from randomly $[0.0,0.2,0.5,0.7]$.
          \item The memory size of this module is chosen randomly from the set $[64,128,256,300,400]$.
          
      \end{itemize}

    \end{enumerate}
    
    \item \textbf{Optimizer} After some trial and error, we found that the model works best for an Adam optimizer\cite{kingma2014adam} initialized with $\beta_1= 0.9, \beta_2 = 0.98$ and $\epsilon = 10^{\minus 9}$. The learning rate was varied by the formula learning\_rate = $d^{−0.5}_{model}$ * min(step\_num$^{\minus 0.5},$step\_num * warmup\_steps$^{\minus 1.5})$. The optimizer and the scheduler is identical to the one chosen in \cite{vaswani2017attention}

\end{itemize}

\end{document}